\documentclass[conference]{IEEEtran}
\IEEEoverridecommandlockouts

\usepackage{amsmath,amssymb,amsfonts}
\usepackage{booktabs}
\usepackage{graphicx}
\usepackage{textcomp}
\usepackage{natbib}
\usepackage{hyperref}
\usepackage[utf8]{inputenc}

\usepackage[table,xcdraw]{xcolor}
\usepackage[ruled,linesnumbered]{algorithm2e}
\def\BibTeX{{\rm B\kern-.05em{\sc i\kern-.025em b}\kern-.08em
    T\kern-.1667em\lower.7ex\hbox{E}\kern-.125emX}}
\begin{document}\pagenumbering{arabic}

\title{Capsule Network based Contrastive Learning of Unsupervised Visual Representations\\
}

\author{\IEEEauthorblockN{Harsh Panwar}
\IEEEauthorblockA{\textit{School of Electronic Engineering and Computer Science} \\
\textit{Queen Mary, University of London}\\
London, UK \\
ec21226@qmul.ac.uk\\
ORCID: 0000-0003-2248-618X}
\and
\IEEEauthorblockN{Ioannis Patras}
\IEEEauthorblockA{\textit{School of Electronic Engineering and Computer Science} \\
\textit{Queen Mary, University of London}\\
London, UK \\
i.patras@qmul.ac.uk \\ ORCID:0000-0003-3913-4738}
}

\maketitle

\begin{abstract}
Capsule Networks have shown tremendous advancement in the past decade, outperforming the traditional CNNs in various task due to it's equivariant properties. With the use of vector I/O which provides information of both magnitude and direction of an object or it's part, there lies an enormous possibility of using Capsule Networks in unsupervised learning environment for visual representation tasks such as multi class image classification. In this paper, we propose Contrastive Capsule (CoCa) Model which is a Siamese style Capsule Network using Contrastive loss with our novel architecture, training and testing algorithm. We evaluate the model on unsupervised image classification CIFAR-10 dataset and achieve a top-1 test accuracy of 70.50\% and top-5 test accuracy of 98.10\%. Due to our efficient architecture our model has 31 times less parameters and 71 times less FLOPs than the current SOTA in both supervised and unsupervised learning. Code available at \url{https://github.com/Harsh9524/CoCa}

\end{abstract}


\section{Introduction}

Unsupervised learning in visual computing is used to learn image representations from pixels without using any annotated labels. The larger part of day-to-day visual information perceived by human beings is in a unsupervised setting without the help of any other form of information \citep{barlow1989unsupervised}. Although much of the work done in the field of artificial intelligence (AI) has been supervised learning, it's time that unsupervised learning is taken into consideration to explore the true potential of AI. \par
Convlutional neural networks (CNN) in the last 10 years has proved to be the most impactful discoveries in the domain of Artificial Intelligence with applications in the Healthcare \citep{panwar2020application}, Education \citep{bhardwaj2021application}, Waste Management \citep{panwar2020aquavision} and many others sectors. But nevertheless CNN comes with it's own disadvantages \citep{gupta2022covid}. CNNs are unable to keep the precise encoding of the location of different objects in an image. This is because of the translation invariant features extracted from the max-pooling layers \citep{mazzia2021efficient}. Additional to this CNNs are not invariant to rotation, reflection, scaling and other affine transformations. To tackle with the issues CNNs proposes there has been a lot of work done in the past years \citep{he2016deep} \citep{simonyan2014very}. The general solution in most cases have been to increase the number of feature maps so that there are enough detectors for all affine transformations. Data augmentation is other techniques which is used to learn the same image in different transformations. Both of these techniques adds redundancy to the architecture and the dataset making it difficult to learn the image representations in a unsupervised way and make predictions. \par
\begin{figure*}[ht]
\centerline{\includegraphics[width=16cm,height=8cm]{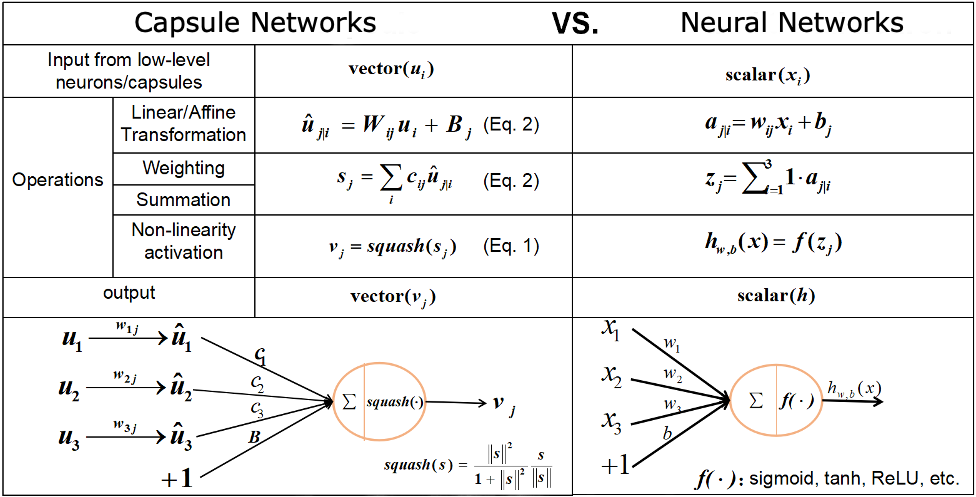}}
\caption{A simple tabular comparison between the basic elements that makes up Capsule Networks and Neural Networks.}
\label{capsulevsneuron}
\end{figure*}

Capsule Networks \citep{sabour2017dynamic} was proposed in 2017 to overcome the drawbacks of CNNs by making use of capsules \citep{hinton2011transforming}, a group of neurons which corporate inside the capsules where each activations inside represent different properties of the same object rather than presence of a specific feature. As seen in fig.~\ref{capsulevsneuron} the inputs from the low level capsules to the higher level capsule are in the form of vectors compared to scalar inputs in traditional neural networks. The lower level capsules represents parts of the whole like nose, eyes and mouth which are sent as vectors to the higher level capsules in form of vector (having direction and magnitude both) where routing by agreement is used to combine to more higher level representation like a face. Because of the transformations "equivariant" properties of capsule networks it holds the potential of being the best option for unsupervised learning. \par

In this paper, we propose Contrastive Capsule (CoCa) Model which uses capsule networks with an updated architecture and contrastive loss function which relies on the instance-wise contrastive learning aiming to bring similar images closer and dissimilar images further away and hence removing the use of annotated labels from the dataset. More particularly, we investigate weather capsule networks can be utilized in siamese networks for unsupervised learning. To the authors best knowledge, this is the first time capsule networks has been investigated in such settings. In addition since the capsule networks are used to replace the CNNs in siamese style contrastive learning methods the number of parameters are reduced by 31 times reducing the training time by 4 times. \par

The main contributions of this paper are:
\begin{enumerate}
    \item We propose Contrastive Capsule (CoCa) Model which achieves 70.50\% top-1 test accuracy and 98.10\% top-5 test accuracy on CIFAR-10 unsupervised image classification task and outperforming the baseline model trained on supervised learning. 
    \item Reduce the total number of parameters by 31 times from 24,620,000 to 780,000 and the FLOPs by 71 times helping the model to train 4 times faster on the same GPU.
    \item Investigate the performance of capsule networks in unsupervised settings and motivate research in this direction.
    \item Investigate the performance of capsule networks with contrastive loss (NT-Xent) in comparison to majorly used margin loss and motivate research in this direction. 
    \item Investigate the compatibility of capsule networks in a siamese architecture and motivate research in this direction. 
\end{enumerate}

The rest of the paper is structured as: Section~\ref{background} gives a brief background and literature survey of significant technologies used in this study, Section~\ref{Methodology} explains the dataset used, architecture proposed and the training algorithm, Section~\ref{Evaluation} evaluates the results obtained, Section~\ref{Discussion} discusses the results and talks about possible improvements and current applications and finally, Section~\ref{Conclusion} concludes the paper. 

\section{Background} \label{background}

We now discuss the background of all the methods use in this paper. Both Capsule Networks and Contrastive Learning are considered state-of-the-art (SOTA) for image classification and unsupervised learning, respectively.

\subsection{Capsule Networks}

Capsule Networks came to existence in 2017 when \cite{sabour2017dynamic} in their paper proposed a unique way of routing between capsules called dynamic routing. But the earliest discussion of capsules can be dated back to 2011 in the paper \cite{hinton2011transforming} which explains how ANNs can learn to convert pixel intensities to pose vectors. Capsules Networks using Dynamic Routing implemented between Primary Capsules and Digit Capsule on MNIST \citep{lecun1998gradient} dataset \cite{sabour2017dynamic} performed well with a test error of 0.25\% with only 3 layers. This was revolutionary results in a way that although similar results have previously been published but only in deep networks and not a shallow network with 3 layers. \par
Since 2017, there has been significant work to update the vanilla capsule networks to scale the performance on bigger datasets. \cite{hinton2011transforming} uses Expectation-Maximisation (EM) routing instead of the traditional dynamic routing and achieve 45\% better results based on test errors compared to the SOTA. Addition to better performance, EM routing Capsule Network shows resistance against adversarial attacks. \cite{wang2018optimization} tackled the routing problem as an optimization problem and in an unsupervised grouping task performed better based on Adjusted Mutual Information (AMI) score (0.914 compared to 0.879). \cite{lenssen2018group} proposes group equivariant capsule networks

\subsection{Contrastive Learning}

Contrastive Learning mainly focuses on Contrastive Loss between the output of a siamese neural network. Before understanding the contrastive loss we need to learn about circle loss and triplet loss. 
Circle Loss \citep{wen2016discriminative} was introduced to handle the drawbacks of Softmax loss function and work on the intuition that the similarity score should be emphasized if it deviates largely from the optimum. Each similarity is re-weighted to showcase the less optimized similarity scores. The loss is measured using the following equation~\ref{eq0.1} as $L_c$
\begin{equation}
\label{eq0.1}
    L_{c} = -\sum_{i=1}^{m} \ln{\frac{exp(W_{y_{i}}^{T}x_{i} + b_{y_{i}} ) }{\sum_{j=1}^{n} exp(W_{j}^{T}x_{i} +b_{j}) }} + \frac{\lambda}{2}\sum_{i=1}^{m} \lVert x_{i} - c_{y_{i}}  \lVert_{2}^{2}
\end{equation}
The disadvantage of Circle Loss is that it fails to penalize inter-class variations and is only useful for intra-class variations. Also calculating the circle loss becomes expensive as the number of classes increases. 
Triplet loss \citep{schroff2015facenet} was introduced where the negative pairs should be distant from the positive pairs by a minimum of margin value. The loss can be easily calculated as shown in equation~\ref{eq0.2}
\begin{equation}
\label{eq0.2}
L_{triplet} = max(dist(anchor,pos) - dist(a,neg) +margin,0)
\end{equation}
where, \\
$pos$: is the positive or the similar image to the anchor.\\
$neg$: negative or the dissimilar image from the anchor.\\
$dist$: is used to calculate the distance between $anchor$, $pos$ and $neg$.\\
$margin$: and m is the margin value which is used to keep the negatives far apart.\\
Triplet Loss showed promising results in different applications like Face Verification and Object Tracking. But triplet has big limitations including a need of paired dataset that is hard to build and the penalty on both positives and negatives is constrained to be the same making it inefficient. This is where constrastive loss steps in. \par 
The concept of contrastive learning was first introduced in 2006 by \citep{hadsell2006dimensionality} as a technique for dimensionality reduction. \par
\begin{figure}[ht]
\centerline{\includegraphics[width=9cm]{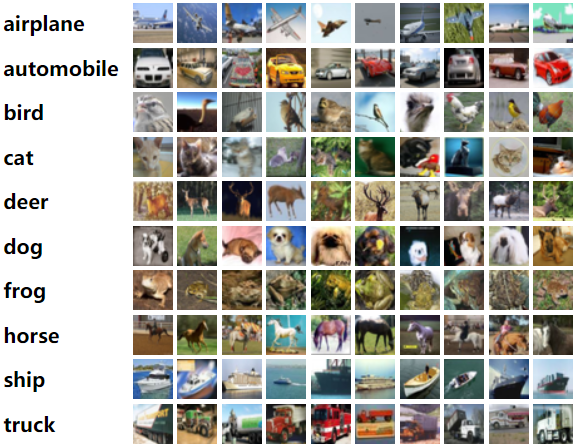}}
\caption{A snapshot of CIFAR-10 dataset comprising of 10 classes.}
\label{Dataset}
\end{figure}

In recent times, contrastive loss is used primarily for semi-supervised and unsupervised learning \citep{chen2020big} \citep{xie2021detco} \citep{koshkina2021contrastive} \citep{Baek_2021_ICCV}. SimCLR presented by \cite{chen2020simple} performed 7\% better than the SOTA based on top-1 accuracy on semi-supervised ImageNet \cite{deng2009imagenet} data. It also matched the performance of a supervised ResNet-50 \cite{he2016deep} and outperforming AlexNet \citep{krizhevsky2012imagenet} on top-5 accuracy with $100\times$ less labels. SimCLRv2 outperforms all the supervised SOTA by only using 10\% of the labels. When 1\% of the labels are used a $10\times$ increase is observed over previous SOTA.


\section{Methodology} \label{Methodology}

\begin{figure}[ht]
\centerline{\includegraphics[width=9cm]{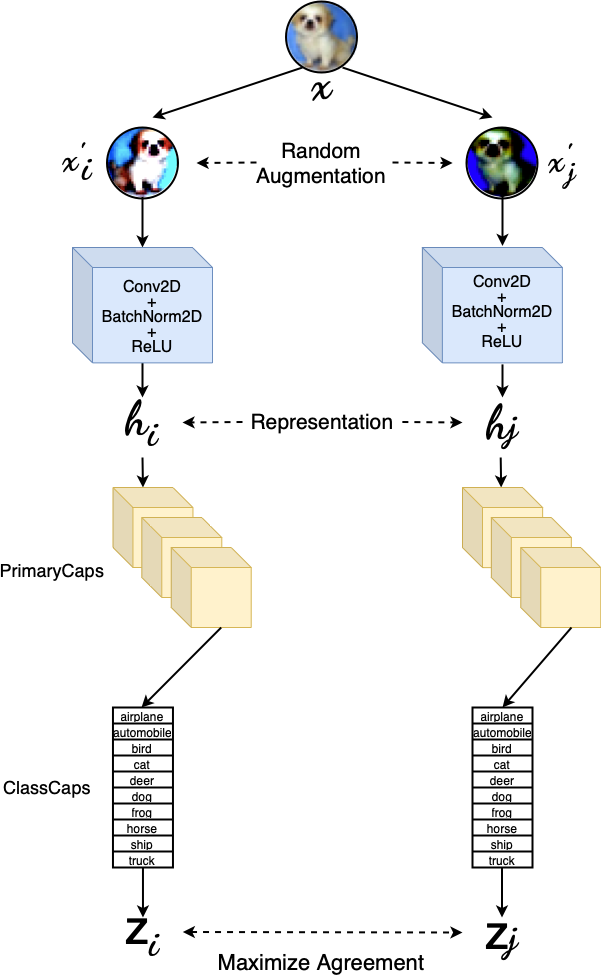}}
\caption{Main architecture of CoCa (proposed) model. }
\label{architecture}
\end{figure}

\subsection{Dataset}
We have used CIFAR-10 \citep{krizhevsky2009learning} dataset on the proposed model as a benchmark. CIFAR-10 60K RBG images of $32\times32$ size each. In total there are 10 classes and each class has 6k images. Each class is mutually exclusive to all other class. For example the automobile class has no overlap of truck or airplane class and so on. The training dataset split comprises of 50,000 sample images while the testing dataset split comprises of 10,000 sample images. 
\subsection{Architecture}

Contrastive Learning requires two sets of images to be passed at once so that the contrastive loss can be calculated. To facilitate this process we use Siamese Network \citep{bromley1993signature} type architecture where two different images are passed to similar architecture and the similarity is measured at the end. The propose architecture as seen in the fig.~\ref{architecture} consists of ConvBlock, PrimaryCaps, ClassCaps as the main components.

\begin{table*}[t]
\centering
\caption{Dimensions and number of parameters for the layers in ConvBlock}
\label{table:table1}
\resizebox{\textwidth}{!}{%
\begin{tabular}{lccccc}
\hline
Layer (type) & input channels & \multicolumn{1}{l}{output channels} & \multicolumn{1}{l}{stride} & \multicolumn{1}{l}{Number of Features} & Number of Parameters \\ \hline 
Conv2d-1           & 3  & 16  & 1 & -   & 432    \\
BatchNorm2d + ReLU & -  & -   & - & 16  & 32     \\
Conv2d-2           & 16 & 32  & 2 & -   & 4,608  \\
BatchNorm2d + ReLU & -  & -   & - & 32  & 64     \\
Conv2d-3           & 32 & 32  & 1 & -   & 9216   \\
BatchNorm2d + ReLU & -  & -   & - & 32  & 64     \\
Conv2d-4           & 32 & 64  & 2 & -   & 18,432 \\
BatchNorm2d + ReLU & -  & -   & - & 64  & 128    \\
Conv2d-5           & 64 & 64  & 1 & -   & 36,864 \\
BatchNorm2d + ReLU & -  & -   & - & 64  & 128    \\
Conv2d-6           & 64 & 128 & 2 & -   & 73,728 \\
BatchNorm2d + ReLU & -  & -   & - & 128 & 256    \\ \hline
\end{tabular}%
}
\end{table*}

\textbf{ConvBlock.} The ConvBlock consists of 6 sets of Conv2D Layers with each Conv2D layer followed by a BatchNorm2d layer and a ReLU layer. The dimension of each layer is shown clearly in the Table \ref{table:table1}. Conv2d layer is used simply to apply a 2D convolution operation over the input image. The input channel to Conv2d-1 layer is the input image with 3 channels (RBG Image). The stride is set to 1 for this layer. For all the Conv2d layers padding is set to be 1 and the kernel size is 3. The BatchNorm2d layer is used to apply Batch Normalization \citep{ioffe2015batch} on a 4D input(3D image plus the batch size) to reduce the internal co-variance shift and reduce training time. Finally, ReLU (Rectified Linear Unit) \citep{fukushima1975cognitron} non-linear activation function is added. Conv2d-6 is the last convolutional layer with highest number of parameters as 73,728. The ConvBlock acts as a feature extractor and outputs $h$ which are later used during testing to create a feature bank. \\ 
\textbf{PrimaryCaps.} The pixel intensities that are inputted into the ConvBlock are converted into activities of local feature detectors and used as input for the PrimaryCaps. Since, they are the first capsule layer in a capsule network they are termed as PrimaryCaps. It is a convolutional capsule layer which has 32 channels of 16D capsules compared to 8D in the original \citep{sabour2017dynamic} paper. It can be implemented using Conv2d with input channel as 128, output channels as 512, stride and padding as 1 and kernel size as 3.
\\
\textbf{ClassCaps.} The final layer is the Class Caps which has an output dimension of 10 each corresponding to one class of CIFAR-10. \\


Compared to the current Deep architectures like ResNet \citep{he2016deep} we can say our architecture is shallow. The total number of parameters it has is 734,800.

\subsection{Algorithm}
The proposed model can be trained using the algorithm presented in \ref{alg:cap} by giving the training data as the input with other hyperparameters --- $\tau$, $W_{ij}$, $b_{mn}$, $\epsilon$ and $\eta$. $\tau$ represents temperature used in the contrastive loss function as shown in eq~\ref{eq1} and is a good hyperparameter to control the strength of penalties if the negative samples are difficult. $W_{ij}$ corresponds to the randomly initialized weight matrix with the shape [num\_capsules, previous\_layer\_nodes, in\_channels, out\_channels]. $\epsilon$ is the number of iteration in the dynamic routing process between the PrimaryCaps and ClassCaps. \par

\SetKwComment{Comment}{/* }{ */}

\begin{algorithm}[!ht]
\caption{CoCa's main training algorithm}
\label{alg:cap}
\SetAlgoLined
\SetKwInOut{Input}{Input}
\SetKwInOut{Output}{Output}
\Input{
\textit{training\_data}: training split from the dataset\\ $\tau$: temperature parameter \\ $W_{ij}$: randomly initialized weight matrix \\ $b_{mn}$: initial log probabilities. \\ $\epsilon$: number of routing  iterations. \\ $\eta$: number of epochs 
} 
\Output{ trained\_model }

\SetKwFunction{FMain}{Capsule Network}
\SetKwProg{Fn}{Function}{:}{}
\Fn{\FMain{$x$, $W_{mn}$}}{
    $h$ $\longleftarrow$ ConvBlock($x$)\;
    $u \longleftarrow$ PrimaryCaps($h$)\;
    $\hat{u}_{mn} = W_{mn}u$\;
    \For{$\epsilon$}{
     $c_{mn} \longleftarrow \frac{exp(b_{mn})}{\sum_{g}exp(b_{mn})}$\;
     $s_n \longleftarrow \sum_{g} c_{gn} u_{gn}$\;
      $y_{n} \longleftarrow \frac{\lVert s_n \lVert s_n}{1 + \lVert s_n \lVert ^2}$\;
      $b_{mn} \longleftarrow b_{mn} + u_{mn} \cdot y_n$ \;
     \KwRet $y_n$}
    \KwRet $y_{n}$, $h$ \;
}

begin\;

\For{$\eta$}{
\For{ $x_i,x_j$ \textbf{in} training\_data }{
 $z_i \longleftarrow$ \texttt{Capsule Network}($x_i$) \;
 $z_j \longleftarrow$ \texttt{Capsule Network}($x_j$)\;
 $z_k \longleftarrow$ CONCAT($z_i$,$z_j$)\;
 $z_k \longleftarrow$ $z_k.transpose()$\;
 $L_{i,j} = - log \frac{exp(sim(z_i,z_j)/\tau_)}{\sum_{k=1}^{2N}exp(sim(z_i,z_k)/\tau)}$ \;
 $L_{i,j}.back\_prop()$\;
 train\_optimizer.$step()$ ;
}
}

\end{algorithm}

\begin{equation}
\label{eq1}
    L_{i,j} = - log \frac{exp(sim(z_i,z_j)/\tau_)}{\sum_{k=1}^{2N}exp(sim(z_i,z_k)/\tau)}
\end{equation}

First, we define \texttt{Capsule Network} function to which the input image and the random initialized weights are passed. As explained in the architecture subsection above the ConvBlock extracts feature representations ($h$) out of the RBG image. The features are then passed to the PrimaryCaps layer which forms an inverse graphics perspective and returns a squashed stack of vectors $u$ using the eq~\ref{eq2}. \par 

\begin{equation}
\label{eq2}
    u_{n} \longleftarrow \frac{\lVert s_n \lVert s_n}{1 + \lVert s_n \lVert ^2}
\end{equation}
Then between the output of PrimaryCaps and between the input of ClassCaps dynamic routing by agreement is executed which helps find the best connections between them and help the two capsule layers communicate. Independent of the type of Image that is inputted into the Network, dynamic routing makes sure that output from PrimaryCaps (child capsule) will be sent to the most relevant ClassCaps (parent capsule).\par

In the initial stage the PrimaryCaps are unsure about which ClassCaps their vector outputs should reach. Coupling co-efficients represented by $c$ is the value that represents the probability of a child capsule output going to a parent capsule and can be seen as a discrete probability distribution across all the ClassCaps. As we can see in step 7 of Algo~\ref{alg:cap} $c$ is calculated by taking the softmax of $b_{mn}$. Then in step 8 we calculate $s_n$ as the sum of the dot product of the input to the ClassCaps $u$ and coupling-coefficient $c$. $s_n$ or the total capsule inputs is squashed using the eq~\ref{eq2} and the output is stored in $y_n$ which is the output of the ClassCaps. In step 10, through the iterative process of dynamic routing by agreement we update the initial log probabilities $b_{mn}$. If the dot product in step 10 between $y_n$ and $\hat{u}_{mn}$ is large then we say that they agree and the coupling coefficient between them increase (since $b_{mn}$ changes) while $c_{mn}$ between other child and parent capsule decreases. The dot product is the agreement measure and it affects how information is weighted in the network. Unlike traditional neural network where we see only feed forward training, here we can observe top-down feedback as well for $\epsilon$ number of iterations. \par
Now that we have defined the \texttt{Capsule Network} function we begin the training part using the for loop in step 14 over the training dataset. $x_i$ and $x_j$ are two images generated from the same image from the training dataset but with different augmentations applied. Both of the images considered as positives are fed to the \texttt{Capsule Network} and we obtain $z_i$ and $z_j$. To generate a hard negative sample we concatenate $z_i$ and $z_j$ and take the transpose of the two tensors and generate $z_k$. Then using eq.~\ref{eq2} in step 19 we calculate the contrastive loss between the positive and negative samples. Back propagation is executed and the model learns through ADAM optimizer. 
\begin{table*}[t]
\centering
\caption{Comparison of state-of-the-art supervised, semi-supervised and unsupervised models on CIFAR-10 dataset}
\label{table:table2}
\resizebox{\textwidth}{!}{%
\begin{tabular}{lccccc}
\hline
\multicolumn{1}{c}{} & \textbf{Method} & \textbf{top-1 test accuracy} & \textbf{top-5 test accuracy} & \textbf{Number of Parameters} & \textbf{FLOPs} \\\hline\\

64 Layered Capsule \citep{xi2017capsule}  & supervised      & 60.54\% & -       & -    & -      \\\\
Baseline CapsNet \citep{sabour2017dynamic}   & supervised      & 68.98\% & -       & 7.9M & -      \\\\
7 Ensemble + 2 Conv \citep{xi2017capsule} & supervised      & 70.50\% & -       & -    & -      \\\\
SimCLR  \citep{chen2020simple}          & semi-supervised & 93\%    & 99\%    & 24.62M  & 1.31G  \\\\
K-means   \citep{zelnik2004self}          & unsupervised    & 22.9\%  & -       & -    & -      \\\\
Sparse AE    \citep{ng2011sparse}       & unsupervised    & 29.7\%  & -       & -    & -      \\\\
GAN  \citep{radford2015unsupervised}               & unsupervised    & 31.5\%  & -       & -    & -      \\\\
DAC       \citep{chang2017deep}          & unsupervised    & 52.2    & -       & -    & -      \\\\
IIC     \citep{ji2019invariant}            & unsupervised    & 61.7\%  & -       & -    & -      \\\\
\hline
\rowcolor[HTML]{C0C0C0} 
CoCa (ours)         & unsupervised    & 70.50\%    & 98.10\% & 780K & 18.34M 
\end{tabular}%
}
\end{table*}


\begin{figure}[ht]
\centerline{\includegraphics[width=9cm]{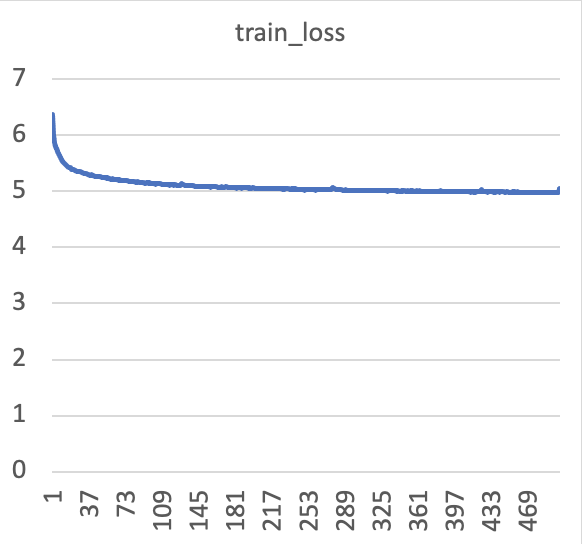}}
\caption{Training loss graph plotted against number of epochs}
\label{train_loss}
\end{figure}

\newcommand{\pluseq}{\mathrel{+}=}
\begin{figure}[ht]
\centerline{\includegraphics[width=9cm]{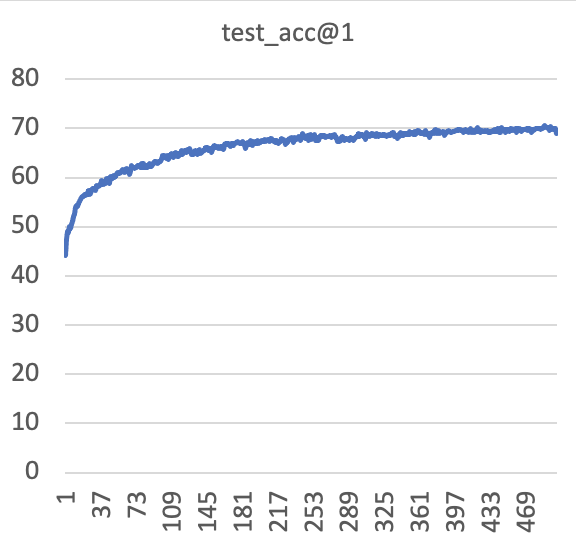}}
\caption{Top-1 test accuracy graph plotted against number of epochs}
\label{test_acc@1}
\end{figure}

\section{Evaluation} \label{Evaluation}

\begin{algorithm}[!ht]
\caption{CoCa's main testing algorithm}
\label{alg:test}
\SetAlgoLined
\SetKwInOut{Input}{Input}
\SetKwInOut{Output}{Output}
\Input{
\textit{memory\_data}: training data split from the dataset without shuffle\\
\textit{test\_data}: testing data split from the dataset\\
$\tau$: temperature parameter\\
$OH_{label}$: a zero matrix\\
$k$: top k most similar images to \\predict the label
}
\Output{ top\_1 test accuracy\\
top\_5 test accuracy
}
begin\;
\For{$x,y$ in $memory\_data$}{
$h,z \longleftarrow$ \texttt{Capsule Network}($x$) \;
$f_{bank} \longleftarrow f_{bank}.append(h)$ \;
$f_{bank} \longleftarrow$ CONCAT($f_{bank}).transpose()$ \;
}
\For{$x,y$ in test\_dataset}{
$f,z \longleftarrow$ \texttt{Capsule Network}($x$) \;
$sim_{matrix} \longleftarrow h \times f_{bank}$ \;
$sim_{weight}, sim_{indices} \longleftarrow  sim_{matrix}.topk(k)$\;
$sim_{weight} \longleftarrow \frac{sim_{weight}}{\tau}$ \;
$OH_{label} \longleftarrow OH_{label}.scatter(index=sim_{labels})$\;
$pred\_scores \longleftarrow \sum OH_{label} * sim_{weight}$\;
$pred\_labels \longleftarrow pred\_scores.argsort()$ \;
$top\_1 \pluseq \sum (pred\_labels[:,:1] \longleftarrow y) $\;
$top\_5 \pluseq \sum (pred\_labels[:,:5] \longleftarrow y) $\;
}
\KwRet $\frac{top\_1}{x.size(0)}\times100, \frac{top\_5}{x.size(0)}\times100$
\end{algorithm}

The CoCa model is evaluated using the algorithm~\ref{alg:test}. We created a $memory\_data$ loader with the same images as the train dataset but without shuffling it. It also takes as input the $test\_data$, $\tau$ which is the same temperature parameter as used in algorithm~\ref{alg:test} and $OH_{label}$ which is the one hot labels. 

First we extract $x$ which is the image data and $y$ which are the labels for the corresponding features from the $memory\_data$. This is the first time we use labels for evaluation purposes and labels were not used during training making our approach truly unsupervised. We input $x$ into our \texttt{Capsule Network} function given in algorithm~\ref{alg:test} and extract $h$ and $z$.
We generate a feature bank by appending the features extracted from the model's feature representations $h$ as shown in fig~\ref{architecture}. The updated feature bank is concatenated and transposed. 
\par
In the second \textbf{for} loop we extract $x$ and $y$ from the testing dataset which is the test split of CIFAR-10 with shuffling enabled. The similarity matrix $sim_{matrix}$ is calculated between the current feature $h$ and the feature bank $f_{bank}$. Using the top k most similar images to predict the label on $sim_{matrix}$ we calculate similarity weight $sim_{weight}$ and similarity indices $sim_{indices}$.
\par
$sim_{weight}$ is then updated by dividing itself with the temperature $tau$. We then calculate the one hot labels $OH\_label$ which is initialized in the input as a zero matrix. Using the scatter function we update the $OH\_label$ based on $sim_{label}$. Finally, prediction scores $pred\_scores$ is calculated as sum of the matrix multiplication of $OH\_label$ and $sim_{weight}$. We then sort the $pred\_scores$ and store it in $pred\_labels$ which are used to calculate the top-1 test accuracy and top-5 test accuracy. The training loss curve can be seen in fig~\ref{train_loss} which is reduced as the number of epochs increases showing that the training is done correctly and since the loss is not increased even after the number of epochs increase we can say that there was no over fitting.
\begin{figure}[ht]
\centerline{\includegraphics[width=9cm]{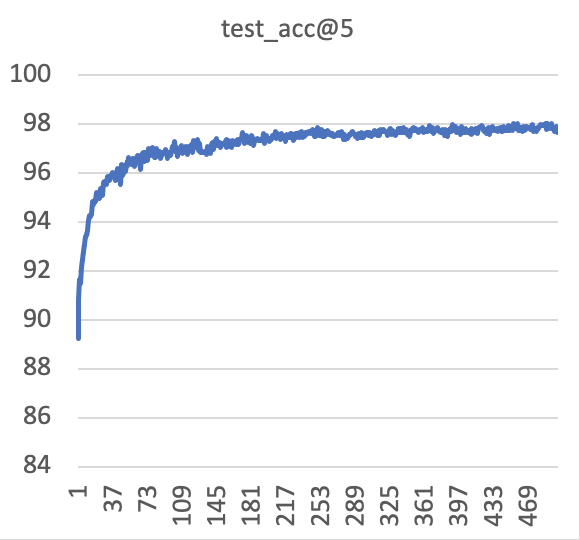}}
\caption{Top-5 test accuracy graph plotted against number of epochs}
\label{test_acc@5}
\end{figure}
Based on the prediction score the top-1 test accuracy is shown in fig~\ref{test_acc@1} and top-5 accuracy shown in fig~\ref{test_acc@5} is calculated. Both the accuracies increase as the number of epoch increase.  \par
The model was trained on PyTorch on a Tesla P100-PCIE-16GB GPU on 500 epochs. The optimized $\tau$ value was found out to be 0.2 after experimenting between 0.1 and 1. For optimization we used ADAM \citep{kingma2014adam} with learning rate = 1e-3 and weight decay 1e-6. The value for $\epsilon$ was taken after careful experimentation to be 3. The final top-1 test accuracy is calculated to be 70.50\% and the top-5 test accuracy is 98.10\% as shown in Table~\ref{table:table2} . The total number of parameters are 780,000 while the number of FLOPs is 18.3 Million.

\section{Discussion} \label{Discussion}

The proposed model CoCa was able to outperform the performance of Supervised Capsule Networks \citep{sabour2017dynamic} on CIFAR-10 without the use of any labels. This sets a big foundation stone for research in the field of unsupervised learning using Capsule Networks. It also shows that use of Contrastive Loss instead of the traditional margin loss for Capsule Networks produce better results. CoCa as compared to vanilla Capsule Networks has 10 times less number of parameters. \par
CoCa achieve a comparable top-5 test accuracy to SimCLR \citep{chen2020simple} while it having 31 times less number of total parameters. CoCa's architecture has a total of 780,000 parameters compared to 24,620,000 which is 31 times less compared to SimCLR \citep{chen2020simple} and 18.34M FLOPs compared to 1.31G FLOPs in SimCLR which is a 71 times reduction in FLOPs. This results in a 4 times decrease in training time on the same GPU and faster execution in real-time scenarios. The top-1 test accuracy of CoCa is Comparatively low to SimCLR's top-1 test accuracy which can be increased keeping in mind the following implementation and architectural changes:\\
\textbf{Routing between Capsules}: The connection between upper level capsules (parent) which represent the whole like an object and lower level capsules (child) which represent part of the whole is known as Routing algorithm. To improve the results of CoCa model we can develop and add an improved routing algorithm. The current advancement in routing algorithms have failed to produce any significant improvements over the baseline dynamic routing with agreement \citep{paik2019capsule}. All the current routing algorithm also suffer through the polarization problem. This is the sole reason that the authors of this paper decided to work with dynamic routing with agreement in the CoCa model in a hope to change the routing algorithm as a future scope. \\
\textbf{Optimization Techniques}:
Although we tried to make the implementation of Contrastive Loss as close as possible to SimCLR's \citep{chen2020big} implementation, there are some optimization techniques we missed including:
\begin{enumerate}
    \item Gaussian blur was not used.
    \item LARS optimizer was not used and instead Adam optimizer with a learning rate of 1e-3 was used.
    \item Linear learning rate scaling was not used
    \item Linear Warmup and CosineLR Schedule was not used. 
\end{enumerate}
All of the above optimization techniques can be used in the future to increase the performance of CoCa model further. \\
\textbf{Larger Batch Size}:
It is noted in the SimCLR \citep{chen2020big} paper that larger batch size results in improved performance of the model. This is because of the large number of negative samples if the batch size is larger in which scenario the contrastive learning is more. Currently due to lack of computation resources we trained our model on a batch size of 512 but increasing it to 8192 can increase the performance of CoCa in future. \\
\textbf{Longer Training Time}:
As per the experiments done in SimCLR \citep{chen2020big} larger number of epochs gives better results in contrastive learning. Again due to computation limitations we trained our model only till 500 epochs but increasing the number of epochs could result in the model learning more and thus increase in test accuracy. 

Exploring unsupervised learning and it's advancement is very essential for the proper use of AI in some of the essential domains such as medical science where the availability of annotations is very limited due to lack of experts \citep{zhang2021self} or mosquito recognition where the number of species is around 3500 \citep{goodwin2021mosquito} while the availability of annotated data is only for 5-10 species. This study can be used as a starting point in this direction.

\section{Conclusion} \label{Conclusion}
This paper introduced the CoCa model, a novel architecture and training algorithm which used Capsule Networks for Contrastive Learning as a technique to learn visual representations in an unsupervised setting. The model outperformed the baseline supervised Capsule Network by achieving a top-1 test accuracy of 70.5\% with 10 times less parameters. It also achieved a comparable top-5 accuracy of 98.10\% with SimCLR but with 31 times less parameters and 71 times reduced FLOPs and hence reducing the training time on same GPU. Through this study we established that Capsule Networks in unsupervised settings is a direction worth exploring and that contrastive loss in a siamese architecture can be one way to do so.

\section{Acknowledgement}
We would like to thank Meta AI for developing the open source PyTorch which is the main framework used to implement CoCa model. We are also thankful to Google for providing cheap GPUs in the form of Google Colaboratory. Finally, we would like to extend our gratitude to
\href{https://github.com/leftthomas}{Hao Ren} and \href{https://github.com/cezannec}{Cezanne Camacho} for their helpful implementations of Capsule Networks and SimCLR, respectively.

\bibliographystyle{agsm}
\bibliography{mybibfile}

\end{document}